\documentclass[letterpaper, 10 pt, conference]{ieeeconf}  

\IEEEoverridecommandlockouts                              

\usepackage[letterpaper, margin=0.8in]{geometry}
\usepackage{graphicx}
\usepackage{booktabs}
\usepackage{color}
\usepackage{times}
\usepackage{amsmath,amssymb,amsfonts}
\usepackage{multirow}
\usepackage{cleveref}
\usepackage{xspace}
\usepackage[table]{xcolor}
\usepackage{pifont}

\newcommand{\real}[1]{\tilde{#1}}

\newcommand{\ba}{\mathbf{a}}
\newcommand{\cmark}{\ding{51}}
\newcommand{\xmark}{\ding{55}}
\title{\LARGE \bf
Social-WM: Safety-Aware Latent World Models for Robot Social Navigation
}

\author{Zhihao Zheng$^{1}$ , Mooi Choo Chuah$^{1}$
\thanks{$^{1}$Computer Science and Engineering department, P.C. Rossin College of Engineering and Applied Science, Lehigh University, Bethlehem, PA 18015, USA. 
        {\tt\small \{zhzc21@lehigh.edu, chuah@cse.lehigh.edu\}}}%
}

\begin{document}

\maketitle
\thispagestyle{empty}
\pagestyle{empty}




\begin{abstract}
Safe social navigation requires a robot to anticipate not only the future consequences of its actions, but also whether a nominal action can actually be executed under surrounding physical and social constraints. We present Social-WM, an efficient latent world-model planning framework trained from egocentric RGB video sequences. Our key observation is that social-navigation experience contains a systematic discrepancy between the \emph{nominal action} and the \emph{realizable action}: a nominal forward action may be fully executed in free space, but needs to be constrained when heading towards a pedestrian or obstacle. Social-WM learns these safety-relevant consequences directly through action-conditioned future prediction, where the target is the actual observed future following each command. We further introduce a realizable inverse-dynamics objective that associates observed latent transitions with the action actually realized rather than the nominal one. At deployment, candidate actions are imagined through the latent world model, and the inverse dynamics model estimates their realizability; nominal--realizable discrepancy then provides a safety signal before execution. The learned dynamics and realizability model remain goal-independent and support both position- and image-goal navigation. On Social-HM3D, Social-WM achieves 63.77\% success while reducing human collisions to 21.67\%, and maintains strong performance under zero-shot transfer to Social-MP3D, without explicit pedestrian tracking, privileged human state, or online reinforcement learning.
\end{abstract}    
\section{Introduction}
\label{sec:intro}

Robots operating in human environments must reach their goals while avoiding collisions and respecting the personal space of nearby pedestrians~\cite{mavrogiannis2023core}. Unlike navigation in static scenes, social navigation involves coupled dynamics: robot motion affects nearby people, while pedestrians and scene geometry constrain which robot actions can be safely executed. Effective navigation therefore requires not only goal-directed control, but also the ability to anticipate the consequences of candidate actions before execution.

World models provide a natural mechanism for such foresight by predicting how the environment evolves under actions. Recent social-navigation methods increasingly incorporate future reasoning. Falcon~\cite{gong2025cognition} predicts future pedestrian trajectories for socially aware policy learning, while NavThinker~\cite{hu2026navthinker} couples an action-conditioned world model with a DD-PPO policy by predicting future scene and interaction representations. These results demonstrate the value of future prediction, but leave an important question underexplored: \emph{can a world model also reason about whether a nominal action is actually realizable under the physical and social constraints?}

\begin{figure}[!t]
    \centering
    \includegraphics[width=0.45\textwidth]{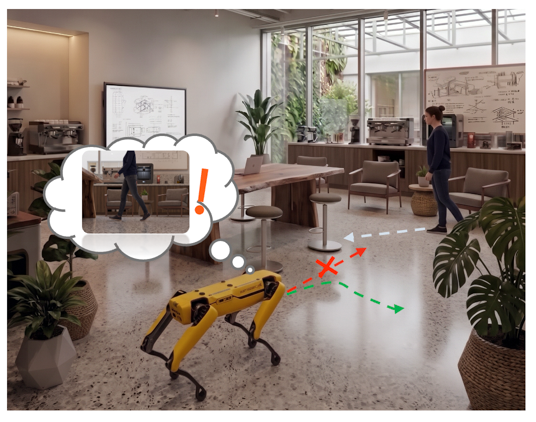}
    \vspace{-2mm}
    \caption{\textbf{Motivation of Social-WM.} A nominal action that appears feasible from the current observation may become unsafe when executed. Social-WM predicts its future consequence and the corresponding realizable action before execution, allowing the planner to reject unsafe candidates and select a safer alternative.}
    \label{fig:motivation}
    \vspace{-2mm}
\end{figure}

This distinction is important in social navigation. The same nominal forward action may be fully executed in free space, but needs to be constrained when heading towards a pedestrian or obstacle. We therefore distinguish the \emph{nominal action} from the \emph{realizable action}, i.e., the motion that can actually occur under the current physical and social constraints.

We introduce \textbf{Social-WM}, an efficient latent world-model framework that jointly learns future dynamics and action realizability from egocentric navigation experience. 
Because supervision comes from the observation that actually follows the command, the model is naturally exposed to outcomes such as normal progress, blocked motion, and constrained execution. 
We further introduce a \emph{realizable inverse-dynamics} objective that requires each observed latent transition to recover the motion actually realized by the robot rather than the issued command, thereby explicitly grounding the learned dynamics in physically achievable motion.


At deployment, Social-WM uses these learned dynamics directly for planning. A lightweight trajectory generator proposes multiple action chunks, which are imagined through the world model. The inverse-dynamics model then estimates the realizable action associated with each proposed transition. A large discrepancy between the nominal and realizable actions indicates that a candidate is likely to be blocked or otherwise inadmissible. Social-WM therefore performs closed-loop \emph{propose--imagine--evaluate--select} planning using both goal progress and action realizability. Because the dynamics and realizability models are goal-independent, the same world model supports both position-goal and image-goal navigation.

We evaluate Social-WM on Social-HM3D and under zero-shot transfer to Social-MP3D.
Social-WM achieves competitive navigation success while substantially reducing human collisions and improving personal-space compliance, using only RGB observations and offline training without explicit pedestrian tracking, privileged human state, or online reinforcement learning.

Our contributions are threefold:
\begin{itemize}[leftmargin=1.2em]
    \item We formulate the discrepancy between nominal and realizable actions as a safety-relevant signal for learning action-conditioned latent dynamics in social navigation.
    \item We introduce realizable inverse dynamics, which associates latent transitions with actually achievable motion and identifies admissible candidate actions from imagined futures before execution.
    \item We develop an efficient goal-independent latent world-model planner that supports both position- and image-goal social navigation and generalizes from Social-HM3D to Social-MP3D.
\end{itemize}

\section{Related Work}
\label{sec:related}

\begin{figure*}[!t]
    \centering
    \includegraphics[width=\textwidth]{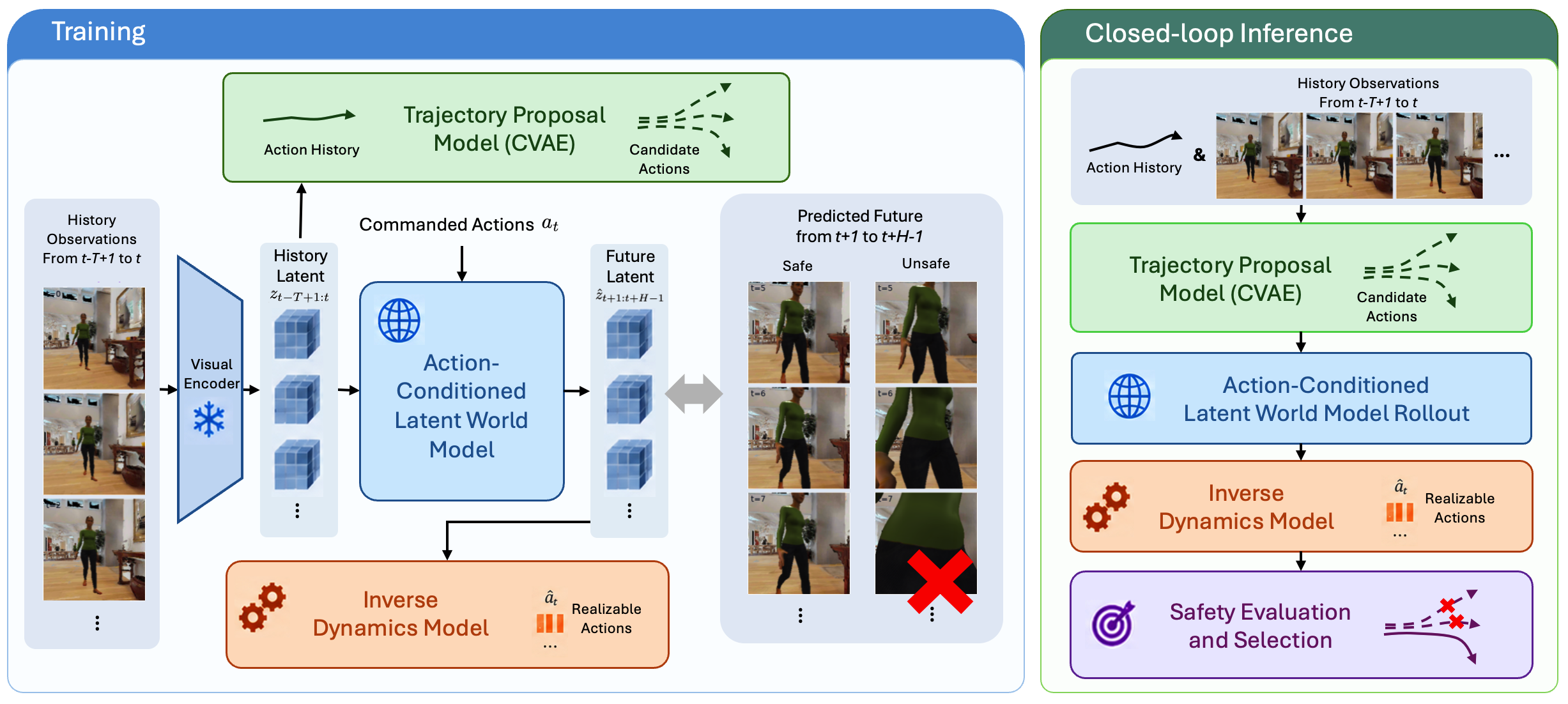}
    \vspace{-4mm}
    \caption{\textbf{Overview of Social-WM.} Candidate action chunks are proposed and imagined through the latent world model. Realizable-action inverse dynamics estimates the motion associated with each imagined transition, enabling candidate selection based on both goal progress and action realizability.}
    
    \label{Fig:framework}
    \vspace{-2mm}
\end{figure*}

\subsection{Social Robot Navigation}

Social navigation requires robots to reach goals while maintaining safety and
comfort around nearby people~\cite{mavrogiannis2023core,francis2025principles}.
Classical methods include social-force models~\cite{helbing1995social} and
reciprocal collision avoidance~\cite{van2011reciprocal}, while more recent
approaches incorporate predictions of future occupancy or pedestrian
motion~\cite{thomas2023foreseeable,poddar2023crowd}. Tightly coupled methods
further model interactions between robot actions and crowd dynamics, e.g.,
through bilevel MPC~\cite{samavi2024sicnav} or learned transition
~\cite{cui2021learning}.

Falcon~\cite{gong2025cognition} predicts future human trajectories to improve
social navigation, while NavThinker~\cite{hu2026navthinker} introduces an
action-conditioned world model whose imagined features are fused into a
DD-PPO policy. Social-WM instead learns from offline RGB trajectories and uses
imagined latent transitions directly for continuous action selection. Rather
than predicting pedestrians explicitly or using future features only as policy
inputs, we evaluate whether a commanded action is consistent with the motion
that is actually realizable under physical and social constraints.

\subsection{World Models for Embodied Decision-Making}

World models learn action-conditioned dynamics for planning or policy
learning. PlaNet~\cite{hafner2019learning} plans in a learned latent space, while
DreamerV2~\cite{hafner2020mastering} optimizes behavior through imagined
trajectories. More recent methods predict directly in visual representation
space. DINO-WM~\cite{zhou2024dino} predicts pretrained visual features for
zero-shot planning, while LeWM~\cite{maes2026leworldmodel} learns compact joint-embedding
dynamics directly from pixels.

Social-WM follows this efficient latent-prediction paradigm but introduces an
explicit notion of action realizability. Compared with
NavThinker~\cite{hu2026navthinker}, which uses imagined features to support an
online-trained policy, our model uses imagined transitions directly to score
candidate action chunks and rejects candidates whose commanded motion is
inconsistent with the predicted realizable motion.

\subsection{Inverse Dynamics and Multimodal Action Generation}

Inverse dynamics predicts the action relating consecutive states and is widely
used to learn control-relevant representations
\cite{agrawal2016learning,pathak2017curiosity}. Standard formulations associate an
observed transition with the recorded action. In social navigation, however,
the issued command may differ from the resulting motion because of obstacles
or nearby pedestrians. We therefore supervise inverse dynamics using
odometry-derived \emph{realizable motion} and apply the same model to imagined
transitions at inference. The discrepancy between commanded and predicted
realizable actions then provides a safety signal for candidate selection.

\section{Methodology}
\label{sec:method}

\subsection{Problem Formulation}
\label{sec:problem}

We consider egocentric navigation in indoor environments populated by $N$ moving humans. The robot observes the scene through a single forward-facing RGB camera and therefore operates under partial observability due to occlusion and a limited field of view. Following~\cite{hu2026navthinker}, we denote the underlying state as
\begin{equation}
s_t \triangleq \big(x_t,\;\mathcal{M},\;\{h_t^{(i)}\}_{i=1}^{N}\big),
\label{eq:state}
\end{equation}
where $x_t$ is the robot pose, $\mathcal{M}$ denotes the static environment, and $h_t^{(i)}$ is the state of the $i$-th human. Neither $\mathcal{M}$ nor the human states are directly observed. At each step, the robot receives an RGB observation $I_t$, its own pose $x_t$, and a goal specification $g$, instantiated as either a goal image or a target position in Sec.~\ref{sec:select}. The robot executes continuous body-frame actions $a_t=(a_t^{\mathrm{fwd}},a_t^{\mathrm{lat}},a_t^{\mathrm{yaw}})\in\mathbb{R}^3$ and plans over $H$-step action chunks $\ba_{t:t+H-1}\in\mathbb{R}^{H\times3}$.


A key property of social navigation is that the nominal action is not
always the action that can be safely executed. We denote the issued command
by $a_t$ and define the \emph{realizable action} $\tilde{a}_t$, aligned with the pose change between $x_t$ and $x_{t+1}$. Importantly, $\tilde{a}_t$ reflects the physical and social safety constraints in the environment.
In free space, the nominal and realizable actions are typically consistent, i.e., $a_t \approx \tilde{a}_t$. When the nominal action conflicts with an obstacle or nearby pedestrian, the resulting motion may be constrained, leading to $a_t \neq \tilde{a}_t$. We therefore treat the
nominal--realizable discrepancy as a safety-relevant signal rather than
execution noise.

\subsection{Overview}
\label{sec:overview}

Social-WM builds on the compact latent world-model design of LeWM~\cite{maes2026leworldmodel} to enable efficient closed-loop planning in dynamic social environments. Our key extension is that our world model is trained from transitions that already contain the realizable consequences of safety constraints: given the current visual state and a nominal action, its prediction target is the RGB observation that actually follows. As a result, the latent dynamics can learn that a nominal command needs to be stopped or altered when it leads to unsafe outcomes.

We jointly ground these future predictions using realizable inverse dynamics. The same transition used to supervise future prediction is also required to recover the action that was realizable rather than the nominal ones. During training, the inverse dynamic model is applied to latent transitions, explicitly aligning with realizable futures under safety constrains. At deployment, candidate action chunks are proposed from the latent observation history and rolled forward through the world model. Their imagined futures are evaluated for both goal progress and action realizability, allowing Social-WM to use the learned future dynamics directly for closed-loop social navigation planning.

\subsection{Safety-Aware Latent World Model}
\label{sec:lewm}

\paragraph{Visual representation}
A frozen DINOv2-small~\cite{oquab2023dinov2} encodes each RGB observation $I_t$ into spatial patch tokens. We discard the class token and adaptively pool the patch grid to $M$ tokens, followed by a lightweight projector that produces the latent representation $z_t\in\mathbb{R}^{M\times D}$, $D$ is feature dimension. 

\paragraph{Action-conditioned future prediction.}
Given the latent history and the corresponding nominal actions, a transformer predictor $F_\phi$ predicts the next latent feature,
\begin{equation}
\hat{z}_{t+1}
=
F_{\phi}\!\left(
z_{t-T+1:t},
a_{t-T+1:t}
\right).
\label{eq:predict}
\end{equation}
Crucially, although the predictor is conditioned on the \emph{nominal} actions, its target $z_{t+1}=E(I_{t+1})$ represents the \emph{actual observed future}. Therefore, a command that is refused is paired with the resulting stopped or constrained future rather than a hypothetical future where the command was fully executed. This allows the world model to directly learn the
safety-relevant consequences from the real-world transitions.

The model predicts entirely in latent space without reconstructing RGB images or explicitly estimating depth, pedestrian trajectories, or human states. Following LeWM~\cite{maes2026leworldmodel}, We train with one-step teacher-forced objective and a representation regularizer objective, as well as an additional multi-step open-loop objective:
\begin{equation}
\mathcal{L}_{\mathrm{pred}}
=
\lambda_{\mathrm{tf}}\mathcal{L}_{\mathrm{tf}}
+
\lambda_{\mathrm{ol}}\mathcal{L}_{\mathrm{ol}}
+
\lambda_{\mathrm{sig}}\mathcal{L}_{\mathrm{sig}}.
\label{eq:predloss}
\end{equation}
The multi-step training objective is particularly important for reducing the accumulated prediction error during the planner's multi-step rollouts.

\begin{figure*}[!t]
    \centering
        \includegraphics[width=0.95\textwidth]{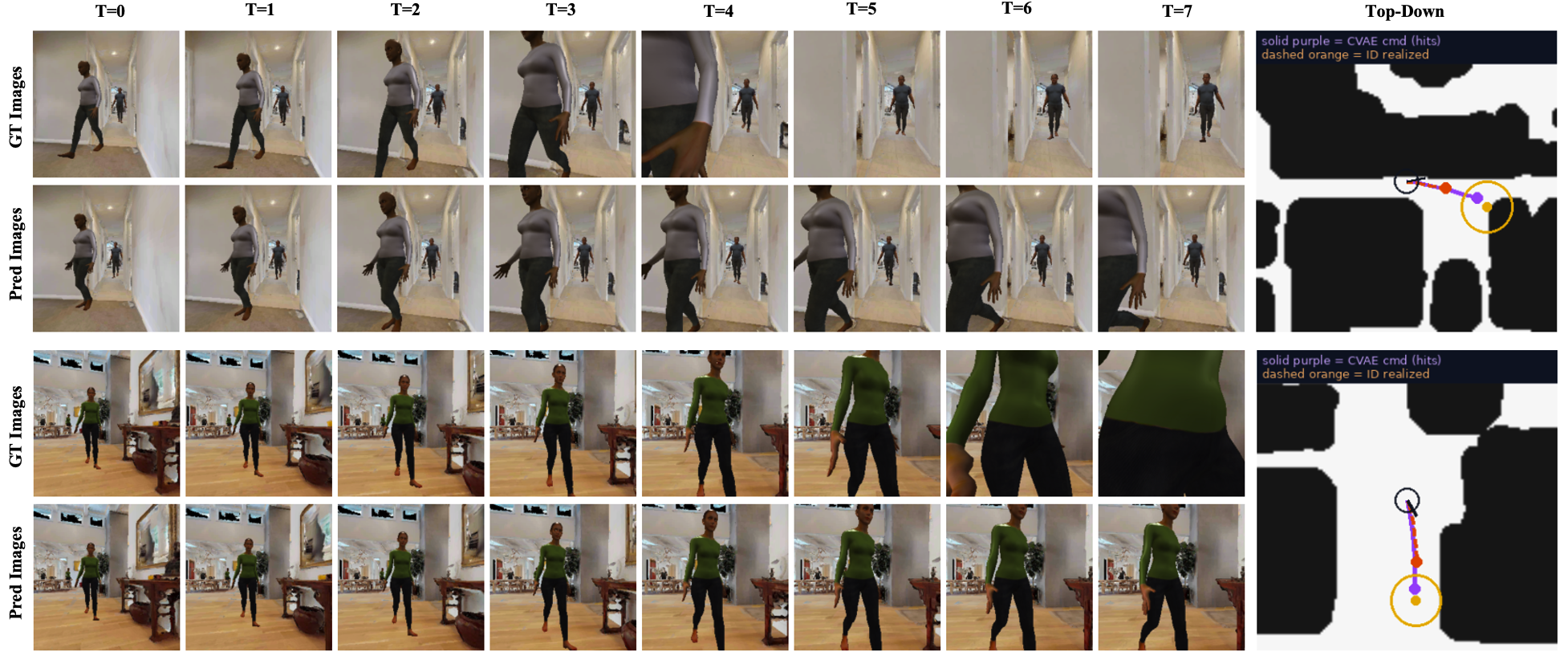}
    \vspace{-2mm}
\caption{\textbf{Qualitative visualization of nominal and realizable futures.}
For each example, the top row shows the ground-truth observations obtained from
the action chunk proposed by the CVAE, while the bottom row visualizes the
future predicted by Social-WM. In the top-down view, the solid purple trajectory
shows the proposed nominal motion and the dashed orange trajectory shows the
ID-predicted realizable motion. The small yellow dot marks the pedestrian
position, and the surrounding large yellow circle denotes its safety
region; the black circle marks the robot's starting position.
When a nominal action conflicts with a pedestrian, Social-WM predicts a
constrained future and the inferred realizable motion deviates from the nominal
motion, exposing the unsafe candidate before execution.}
\label{fig:qualitative}
    \vspace{-3mm}
\end{figure*}

\subsection{Realizable Inverse Dynamics}
\label{sec:id}

The future prediction objective already exposes the world model to the actual consequences of nominal actions, including transitions where an action is blocked or refused. However, latent prediction alone does not explicitly indicate how much of the nominal action was realizable. We therefore introduce a realizable inverse-dynamics objective that associates observed latent transitions with the action the robot actually executes.

Similar to INTACT~\cite{sun2026intact}, we use an inverse-dynamics model $I_\psi$ with a local intent $m_t^{\ell}=z_{t+1}-z_t$ to predict the corresponding realizable action,
\begin{equation}
\hat{\real{a}}_t = I_\psi\!\left(z_t,\,m_t^{\ell}\right)  
\label{eq:id}
\end{equation}
Unlike conventional inverse dynamics that recovers the demonstrated command, we supervise the model using the realizable action $\real{a}_t$ recovered from consecutive robot poses:
\begin{equation}
\mathcal{L}_{\mathrm{id}}
=
-\log p_\psi\!\left(
\real{a}_t
\mid
z_t,\,m_t^{\ell}
\right)
\label{eq:idloss}
\end{equation}
This distinction is important because the nominal action $a_t$ alone does not reveal whether it was fully executed, constrained, or refused. Predicting $\real{a}_t$ instead requires the latent transition to preserve the scene information associated with safety.

Together with the latent prediction objectives, the complete training objective is
\begin{equation}
    \mathcal{L}_{\mathrm{WM}} =
    \lambda_{\mathrm{tf}}\mathcal{L}_{\mathrm{tf}}
    +\lambda_{\mathrm{ol}}\mathcal{L}_{\mathrm{ol}}
    +\lambda_{\mathrm{sig}}\mathcal{L}_{\mathrm{sig}} +\lambda_{\mathrm{id}}\mathcal{L}_{\mathrm{id}}
    \label{eq:wm}
\end{equation}
Thus, Social-WM learns complementary information from the same navigation experience: the action-conditioned world model predicts the actual future resulting from a nominal action, while inverse dynamics explicitly associates the resulting latent transition with realizable action. At deployment, the inverse-dynamics model predicts the realizable action from imagined future, allowing the planner to identify potentially unsafe actions before execution.

\begin{table*}[t]
\centering
\caption{
\textbf{Social Navigation Results on Social-HM3D and Social-MP3D}
(zero-shot transfer from Social-HM3D).
We report SR/SPL (success/efficiency), PSC (social compliance), and H-Coll
(human collisions). Best and 2nd-best are in \textbf{bold} and \underline{underline}.
}
\label{tab:main_results}

\setlength{\tabcolsep}{9pt}

\renewcommand{\arraystretch}{1.15}
\resizebox{0.9\textwidth}{!}{
    \begin{tabular}{lcccc|cccc}
    \toprule
    
    \multirow{2}{*}{\textbf{Methods}}  & \multicolumn{4}{c|}{\textbf{Social-HM3D}} & \multicolumn{4}{c}{\textbf{Social-MP3D}} \\
    
    \cmidrule(lr){2-5}
    \cmidrule(lr){6-9}
    
    & \textbf{SR}$\uparrow$ & \textbf{SPL}$\uparrow$ & \textbf{PSC}$\uparrow$ & \textbf{H-Coll}$\downarrow$    & \textbf{SR}$\uparrow$    & \textbf{SPL}$\uparrow$    & \textbf{PSC}$\uparrow$    & \textbf{H-Coll}$\downarrow$ \\
    
    \midrule
    
    \rowcolor{gray!15}\multicolumn{9}{l}{\textbf{Rule-based}} \\
    
    A*~\cite{hart1968formal}   & 44.81    & 43.99    & {90.38}    & 54.80    & 45.67    & \underline{44.69}    & 91.97    & 54.00 \\
    
    ORCA~\cite{van2011reciprocal}    & 37.44    & 32.91    &  \underline{92.23}    & {39.77}    & 38.81    & 34.65    & \textbf{94.03}    & {39.86} \\
    
    \midrule
    
    \rowcolor{gray!15}\multicolumn{9}{l}{\textbf{Reinforcement Learning-based}} \\
    
    Habitat-official~\cite{puig2024habitat} & 38.99  & 33.53    & 90.37    & 55.48    & 37.00    & 31.76& 92.03    & 52.33 \\
    
    Falcon~\cite{gong2025cognition}    & {56.26}    & \underline{52.05}    & 89.76    & 41.22    & \underline{51.67}    &  \textbf{45.54}    & 92.53    & 40.67 \\
    
    {NavThinker}~\cite{hu2026navthinker}    &  \underline{59.46}    & \textbf{55.00}    & 89.91    &  \underline{39.09}    & {47.33}    & 41.71    & {93.68}    &  \underline{37.67} \\
    
    \midrule
     \rowcolor{gray!15}\multicolumn{9}{l}{\textbf{Ours}} \\
     
    {Social-WM}   & \textbf{63.77} & {49.62} & \textbf{94.37} & \textbf{21.67} & \textbf{54.15} & 44.32 & \underline{93.95} & \textbf{23.34}\\
    
    \bottomrule
    \end{tabular}
}
\end{table*}

\subsection{Trajectory Proposal}
\label{sec:propose}

The world model predicts the consequences of candidate actions but does not itself generate them. We therefore use a lightweight generative model $G_\theta$ conditioned on the recent latent and action history to produce $K$ plausible $H$-step action chunks,
\begin{equation}
\ba^{(k)}_{t:t+H-1}
=
G_\theta
\left(
z_{t-T+1:t},
a_{t-T+1:t},
\epsilon^{(k)}
\right),
\quad
k=1,\ldots,K,
\label{eq:gen}
\end{equation}
where $\epsilon^{(k)}$ introduces stochasticity. 

Our default implementation uses a conditional variational autoencoder (CVAE), whose compact stochastic latent enables parallel generation of diverse local behaviors such as proceeding, turning, waiting, or bypassing. The proposal model is goal-independent: its role is to provide plausible alternatives, while goal progress and realizability are evaluated after each candidate is imagined by the world model. The CVAE is trained with the standard reconstruction and KL objectives,
\begin{equation}
\begin{aligned}
\mathcal{L}_{\mathrm{CVAE}}
= \;&
\mathbb{E}_{q(u\mid\ba,c_t)}
\left[
\|\ba-D_\theta(u,c_t)\|_1
\right] \\
 & + \beta D_{\mathrm{KL}}
\left(
q(u\mid\ba,c_t)\|p(u)
\right).
\end{aligned}
\label{eq:cvae}
\end{equation}

\subsection{Safety Evaluation and Selection}
\label{sec:select}

At deployment, each candidate action chunk is rolled through the frozen world model, producing an imagined latent trajectory $\hat{z}^{(k)}_{t+1:t+H}$. The inverse-dynamics model is then applied to each imagined transition to estimate its realizable action before execution. We measure the discrepancy between the nominal and predicted realizable actions as
\begin{equation}
c_{\mathrm{safe}}^{(k)}
=
\frac{1}{H}
\sum_{\tau=t}^{t+H-1}
\left\|
a_\tau^{(k)}
-
I_\psi
\left(
\hat z_\tau^{(k)},
\hat z_{\tau+1}^{(k)}-\hat z_\tau^{(k)}
\right)
\right\|.
\label{eq:safe}
\end{equation}
A small residual indicates that the imagined future is consistent with the nominal actions, whereas a large residual indicates that the candidate is likely to be constrained. For example, if a candidate commands forward actions toward a pedestrian while the world model predicts a stopped future, the inverse-dynamics model estimate the same stopped action and $c_{\mathrm{safe}}^{(k)}$ increases. 

We treat action realizability as a feasibility requirement. Candidates satisfying $c_{\mathrm{safe}}^{(k)}<\tau$ form the admissible set
\begin{equation}
\mathcal{C}^{\mathrm{safe}}_t
=
\left\{
\ba^{(k)}
\mid
c_{\mathrm{safe}}^{(k)}<\tau
\right\},
\label{eq:safeset}
\end{equation}

from which the controller selects the candidate with the lowest goal-conditioned cost,

\begin{equation}
\ba^\star
=
\arg\min_{\ba^{(k)}\in\mathcal{C}^{\mathrm{safe}}_t}
c_{\mathrm{task}}^{(k)}(g).
\label{eq:plan}
\end{equation}
The safety evaluation itself is independent of the goal representation; the goal enters only through $c_{\mathrm{task}}$.

\paragraph{Position Goal}
Our primary benchmark setting provides the final target position $p_g$. Each candidate action chunk is integrated kinematically and ranked by its closest distance to the target,
\begin{equation}
c_{\mathrm{task}}^{(k)}
=
\min_{\tau}
\left\|
\hat{x}^{(k)}_\tau-p_g
\right\|.
\label{eq:poscost}
\end{equation}
The same formulation also accepts an intermediate position subgoal in place of the final target. Since this geometric cost does not account for humans or obstacles, candidates with strong nominal progress may still be rejected by the realizability constraint.

\paragraph{Image Goal}
Social-WM can alternatively receive a goal image $I_g$, encoded by the same frozen visual encoder. Candidates are ranked according to the similarity between their imagined latent and the goal latent,
\begin{equation}
c_{\mathrm{task}}^{(k)}
=
-\max_{\tau\in\mathcal{T}_k}
\cos\!\left(
 z_\tau^{(k)},
 z_g
\right),
\label{eq:imgcost}
\end{equation}
where $\mathcal{T}_k$ contains the rollout steps considered for comparison. We treat the modality and granularity of goal guidance as orthogonal to Social-WM and study image-goal versus position-goal separately in Sec.~\ref{sec:goal_ablation}.

\paragraph{Closed-loop execution}
Only the first action of the selected chunk $\ba^\star$ is executed. The robot then observes the resulting RGB frame and repeats the complete propose--imagine--evaluate--select cycle every step. Candidate generation and latent rollouts are evaluated in parallel, retaining the efficiency of the compact LeWM backbone while providing multi-step foresight for social navigation.
\section{Experiments}
\label{sec:experiments}

\subsection{Experimental Setup}

\noindent\textbf{Benchmarks.}
We evaluate Social-WM on Social-HM3D~\cite{gong2025cognition}, following the single-robot protocol of NavThinker~\cite{hu2026navthinker}. The benchmark uses Habitat ~\cite{puig2024habitat} scenes with multiple goal-driven pedestrians, where the robot must reach its target while avoiding collisions and respecting personal space. We additionally evaluate zero-shot transfer to Social-MP3D without retraining or adaptation.

\noindent\textbf{Metrics.}
Following NavThinker~\cite{hu2026navthinker}, we report Success Rate (SR), Success Weighted by Path Length (SPL), Personal Space Compliance (PSC), and Human Collision rate (H-Coll). SR measures the fraction of episodes in which the robot successfully reaches its goal, while SPL additionally accounts for path efficiency. PSC measures compliance with pedestrian personal space, and H-Coll reports the fraction of episodes involving a collision with a human.

\noindent\textbf{Baselines.}
We compare Social-WM with SOTA methods, including the rule-based planners A*~\cite{hart1968formal} and ORCA~\cite{van2011reciprocal}, the Habitat-official reinforcement-learning baseline~\cite{puig2024habitat}, the future-aware social-navigation method Falcon~\cite{gong2025cognition}, and the action-conditioned world-model method NavThinker~\cite{hu2026navthinker}.

\begin{figure}[!t]
    \centering
        \includegraphics[width=0.45\textwidth]{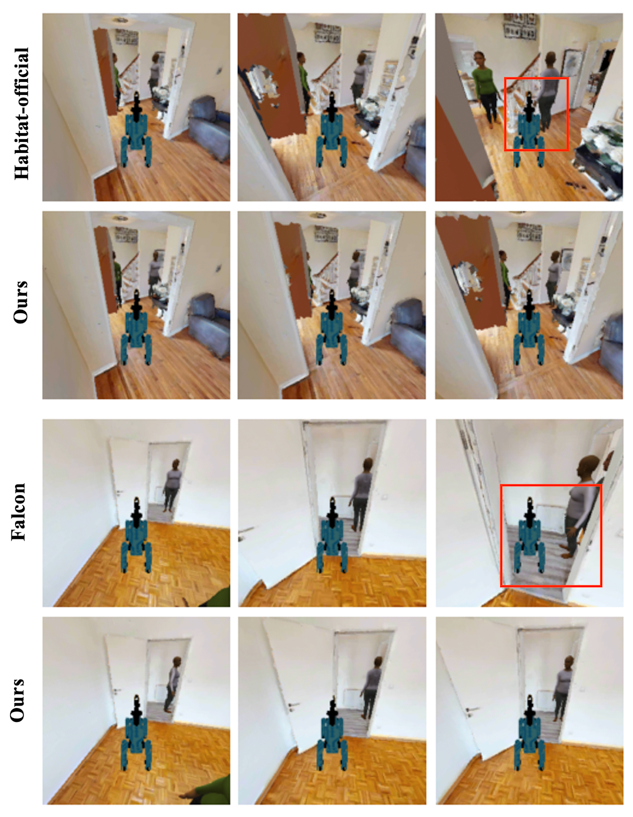}
    \vspace{-2mm}
    \caption{\textbf{Qualitative comparison on Social-HM3D.} Red boxes highlight unsafe interactions with insufficient pedestrian clearance. Social-WM instead slows or waits to maintain enough separation from pedestrians.}

    \label{fig:qualitative_comparison}
    \vspace{-2mm}
\end{figure}

\subsection{Implementation Details}
\label{sec:implementation}


We collect the training trajectories in Habitat using the Social-HM3D
training split from Falcon, covering 200 scenes, 11,676 episodes, and
463,645 transitions. During collection, a keep-out gate rejects nominal
actions that would enter a pedestrian, causing unsafe commands to produce
stopped or constrained realized motion. Such blocked events occur in
18.1\% of transitions, exposing the world model to the effect of social
safety constraints without providing explicit human states or keep-out
signals.

Frames are encoded at $224\times224$ into $M{=}16$ tokens of dimension
$D{=}384$. The predictor is a six-layer transformer trained with history
$T{=}8$ and future $H{=}8$; the planner evaluates and selects from $K{=}8$ candidates. 
The inverse-dynamics head is randomly initialized on a pretrained trunk
and trained at ten times the trunk learning rate. All modules are trained
 with AdamW, the visual encoder is frozen. 

\subsection{Quantitative Results}
\label{sec:quantitative}

Table~\ref{tab:main_results} compares Social-WM with prior methods under the standard position-goal setting. On Social-HM3D, Social-WM achieves the highest SR (63.77\%) and PSC (94.37\%), while reducing H-Coll to 21.67\%, a 44.6\% relative reduction over NavThinker. Under zero-shot transfer to Social-MP3D, Social-WM again achieves the highest SR (54.15\%) and lowest H-Coll (23.34\%), while maintaining competitive SPL (44.32\%) and PSC (93.95\%). These results show that Social-WM substantially improves social safety while preserving strong task success, and that the learned dynamics generalize across unseen environments.

\subsection{Qualitative Results}
\label{sec:qualitative}

Fig.~\ref{fig:qualitative} illustrates the internal prediction of Social-WM. Since the world model operates entirely in latent space, we visualize each predicted latent using its nearest-neighbor observation from the dataset. Although the CVAE may propose an action sequence that approaches a pedestrian, the predicted latent future reflects the scene constraint, and inverse dynamics estimates realizable actions that stop or deviate from the nominal actions. The corresponding top-down views visualize this nominal--realizable discrepancy before execution.

Fig.~\ref{fig:qualitative_comparison} further compares Social-WM's closed-loop behaviors with Habitat-official \cite{puig2024habitat} and Falcon \cite{gong2025cognition} under pedestrian interactions. Both baselines continue goal-directed actions into pedestrian occupied regions, resulting in unsafe proximity or collision, whereas Social-WM slows, waits, or adjusts its action when the intended progress is not safely realizable. Together, these examples show how safety-aware future prediction translates into safer navigation behavior without explicit pedestrian tracking.

\begin{table}[t]
\centering
\caption{\textbf{Ablation of different goal settings on Social-HM3D.}
We compare final-goal and intermediate-subgoal settings under position and image goals. }
\label{tab:goal_ablation}

\renewcommand{\arraystretch}{1.12}
\begin{tabular}{lcccc}
\toprule
\textbf{Goal Guidance} & \textbf{SR}$\uparrow$ & \textbf{SPL}$\uparrow$ & \textbf{PSC}$\uparrow$ & \textbf{H-Coll}$\downarrow$ \\
\midrule
Final Image       & 42.91 & 27.07 & 92.11 & 26.26 \\
Image Subgoal     & {59.52} & {35.23} & {93.16} & {27.04} \\
Final Position    & \underline{63.77} & \underline{49.62} & \textbf{94.37} & \textbf{21.67} \\
Position Subgoal  & \textbf{68.65} & \textbf{58.72} & \underline{93.28} & \underline{22.09} \\
\bottomrule
\end{tabular}
\end{table}

\begin{table}[t]
\centering
\caption{\textbf{Ablation of different planners on Social-HM3D.} }
\vspace{-2mm}
\label{tab:planner_ablation}

\renewcommand{\arraystretch}{1.12}
\resizebox{0.48\textwidth}{!}{
\begin{tabular}{lccccc}
\toprule
\textbf{Planner} & \textbf{SR}$\uparrow$ & \textbf{SPL}$\uparrow$ & \textbf{PSC}$\uparrow$ & \textbf{H-Coll}$\downarrow$ & \textbf{Latency(ms)}$\downarrow$ \\
\midrule
MPC              & {56.08} & {32.72} & {91.34} & {31.33} & 139\\
CVAE             & \textbf{59.52} & {35.23} & {93.16} & \textbf{27.04} & \textbf{51} \\
Diffusion Policy & {58.84} & \textbf{40.15} & \textbf{94.08} & {31.49} & 126 \\ 
\bottomrule
\end{tabular}
}
\vspace{-3mm}
\end{table}

\subsection{Effect of Navigation Goals}
\label{sec:goal_ablation}

Table~\ref{tab:goal_ablation} studies how goal modality and guidance granularity affect Social-WM while keeping the world model, CVAE trajectory proposal, and safety evaluation unchanged. Following the long-horizon navigation setting of~\cite{shen2026efficient}, we construct a topological memory along the reference route and use it to provide intermediate subgoals. At each step, we first identify the robot's current node in the topology and select the closet node along the route as the immediate subgoal. We use its observation for image-subgoal guidance and its corresponding location for position-subgoal guidance.

Intermediate guidance consistently improves navigation success and efficiency. Image subgoals increase SR from 42.91\% to 59.52\% and SPL from 27.07\% to 35.23\%, while position subgoals improve SR from 63.77\% to 68.65\% and SPL from 49.62\% to 58.72\%. In contrast, PSC and H-Coll remain similar across final-goal and subgoal variants, indicating that intermediate guidance primarily improves long-horizon route progress, while Social-WM's local realizability mechanism remains effective for safety. Position guidance consistently outperforms image guidance, especially without intermediate subgoals, reflecting the greater difficulty of directly matching a distant goal image from short-horizon latent rollouts.

\subsection{Effect of Trajectory Planner}
\label{sec:planner_ablation}

Table~\ref{tab:planner_ablation} compares three trajectory planners while keeping the Social-WM dynamics, safety evaluation, and goal settings fixed. We use the image-subgoal setting for this study because it provides a moderate planning difficulty between direct position goal and the more challenging final-image goal. The MPC baseline follows the planning configuration of LeWM~\cite{maes2026leworldmodel}. CVAE achieves the best overall trade-off, with the highest SR (59.52\%), lowest H-Coll (27.04\%), and substantially lower latency than both MPC and diffusion. Diffusion Policy obtains the best SPL (40.15\%) and PSC (94.08\%), but requires higher inference cost, while MPC is both slower and less effective. These results motivate CVAE as our default trajectory planner for efficient closed-loop navigation.

\subsection{Effect of Realizable-Action Learning}
\label{sec:id_ablation}

Table~\ref{tab:id_ablation} isolates the contribution of realizable-action learning. We use the image-subgoal setting with the MPC planner for a fair comparison, since the CVAE and diffusion planners would require retraining if the world-model latent space changes. Adding the inverse-dynamics objective improves the world model's representation of realizable future, increasing SR from 51.33\% to 53.08\% and SPL from 26.49\% to 30.16\%, while reducing H-Coll from 44.97\% to 42.12\%. Using the learned nominal--realizable discrepancy explicitly for safety evaluation provides a larger gain, further reducing H-Coll to 31.33\% and increasing SR to 56.08\%. Overall, the full Social-WM reduces human collisions by 13.64\% over the latent-WM baseline, showing that realizable-action supervision and its use during planning provide complementary benefits.

To evaluate whether the nominal--realizable discrepancy actually captures safety constraints, we perform a 
transition-level diagnostic. For each observed transition $(z_t,z_{t+1})$,
the inverse-dynamics model predicts the realizable action
$\hat{\tilde{a}}_t$, and we compute the residual
$\|a_t-\hat{\tilde{a}}_t\|_{xy}$ between the nominal action and the predicted
realizable action as in Eq.~\ref{eq:safe}. Each transition is labeled as \emph{blocked} or \emph{free}
according to whether the nominal translation was prevented for safety constraints. 
We evaluate 22,512 transitions from 584 held-out episodes, and
additionally report an \emph{Easy} subset with fewer blocked interactions
($\sim$13\%) and a \emph{Hard} subset with more frequent pedestrian conflicts
(36.8\%), compared with 18.1\% in the full dataset.
As shown in Table~\ref{tab:residual_diag}, blocked transitions produce much
larger nominal--realizable residuals than free transitions, yielding an overall
AUROC of 0.961 and similarly strong separation on both Easy and Hard subsets.
The inverse-dynamics model also maintains a low MAE of 0.10 when predicting
realizable actions on free transitions, indicating that it does preserve normal
motion rather than only predicting reduced actions. These results show that
the learned discrepancy is strongly associated with blocked or constrained
execution.

\begin{table}[t]
\centering
\caption{\textbf{Effect of realizable-action learning on Social-HM3D.}}
\vspace{-2mm}
\label{tab:id_ablation}
\setlength{\tabcolsep}{4.8pt}
\renewcommand{\arraystretch}{1.12}
\begin{tabular}{lcc|cccc}
\toprule
\textbf{Variant} &
$\mathcal{L}_{\mathrm{id}}$ &
\textbf{Safety Eval.} &
\textbf{SR}$\uparrow$ &
\textbf{SPL}$\uparrow$ &
\textbf{PSC}$\uparrow$ &
\textbf{H-Coll}$\downarrow$ \\
\midrule
Latent WM        & \xmark & \xmark & 51.33 & 26.49 & 90.26 & 44.97 \\
+ Realizable ID  & \cmark & \xmark & 53.08 & 30.16 & 90.34 & 42.12 \\
Social-WM        & \cmark & \cmark & \textbf{56.08} & \textbf{32.72} & \textbf{91.34} & \textbf{31.33} \\
\bottomrule
\end{tabular}
\vspace{-3mm}
\end{table}

\subsection{Effect of Latent Representation}
\label{sec:latent_ablation}


Table~\ref{tab:latent_ablation} evaluates whether different latent
representations preserve geometric information and support goal ranking using
the same frame pairs. For pose decodability, a lightweight regression probe
takes the latent features of two frames and predicts their ground-truth relative
motion $(\Delta x,\Delta y,\Delta\mathrm{yaw})$ computed from the corresponding
robot poses. We report $R^2$ for each motion component: values near 1 indicate
accurate pose recovery and negative values indicate poor geometric decodability.

For goal ranking, we use the latent-space goal cost to select one candidate
frame for each goal. \emph{Err@Min} is the physical distance between this
selected frame and the goal, so a smaller value means that latent-space
similarity selects a frame that is physically closer to the target.
\emph{Regret} compares this selected distance with the distance of the
geometrically closest candidate frame. It therefore measures the additional
physical error caused by using latent-space ranking instead of an oracle
geometric ranking.

LeWM-CLS and LeWM-Patch both have negative $R^2_x$, indicating poor
forward-motion decodability. In contrast, DINOv2-Patch is the only
representation with positive $R^2$ on all pose dimensions and also achieves
the best goal ranking, reducing Err@Min from 0.659\,m to 0.320\,m and Regret
from 0.422\,m to 0.083\,m compared with LeWM-CLS. These results motivate our
use of frozen DINOv2 spatial tokens in Social-WM.

\begin{table}[t]
\centering
\caption{\textbf{nominal--realizable residual discrepancy.}
Mean residuals are reported for blocked and free transitions.}
\vspace{-2mm}
\label{tab:residual_diag}
\resizebox{0.8\columnwidth}{!}{
\begin{tabular}{lccc}
\toprule
Subset & Blocked Resid. & Free Resid. & AUROC $\uparrow$ \\
\midrule
All      & 0.97 & 0.26 & 0.961 \\
Easy    & 0.99 & 0.09 & 0.988 \\
Hard    & 0.99 & 0.10 & 0.995 \\
\bottomrule
\end{tabular}
}
\end{table}

\begin{table}[t]
\centering
\caption{\textbf{Latent representation ablation.} Pose decodability and latent goal-ranking quality on frame pairs.}
\vspace{-2mm}
\label{tab:latent_ablation}

\setlength{\tabcolsep}{4.2pt}
\renewcommand{\arraystretch}{1.12}

\begin{tabular}{lccc|cc}
\toprule
\textbf{Representation} &
$\mathbf{R^2_{x}}\uparrow$ &
$\mathbf{R^2_{y}}\uparrow$ &
$\mathbf{R^2_{\mathrm{yaw}}}\uparrow$ &
\textbf{Err@Min}$\downarrow$ &
\textbf{Regret}$\downarrow$ \\
\midrule
LeWM-CLS       & -0.36 & -0.00 & 0.37 & 0.659 & 0.422 \\
LeWM-Patch     & -0.41 &  0.15 & \textbf{0.58} & 0.740 & 0.503 \\
DINOv2-CLS     & -0.10 &  0.16 & 0.31 & 0.756 & 0.520 \\
DINOv2-Patch   & \textbf{0.12} & \textbf{0.29} & 0.42 &
\textbf{0.320} & \textbf{0.083} \\
\bottomrule
\end{tabular}
\end{table}

\subsection{Efficiency and Latency}
\label{sec:efficiency}

Social-WM contains 58M parameters, compared with 18M in the original LeWM. Most of this increase comes from replacing the trained ViT encoder with a 22M-parameter frozen DINOv2 encoder; the additional realizable-action ID head and CVAE remain lightweight at 2.4M and 5M parameters, respectively. Despite the larger model size, planning remains efficient because candidate generation and world-model rollouts operate in latent space. On an NVIDIA A5000, CVAE-based planning requires only 51\,ms per step, compared with 126\,ms for Diffusion Policy and 139\,ms for LeWM-style MPC, making our default planner approximately $2.5\times$ and $3\times$ faster, respectively.

\begin{figure}[!t]
    \centering
        \includegraphics[width=0.45\textwidth]{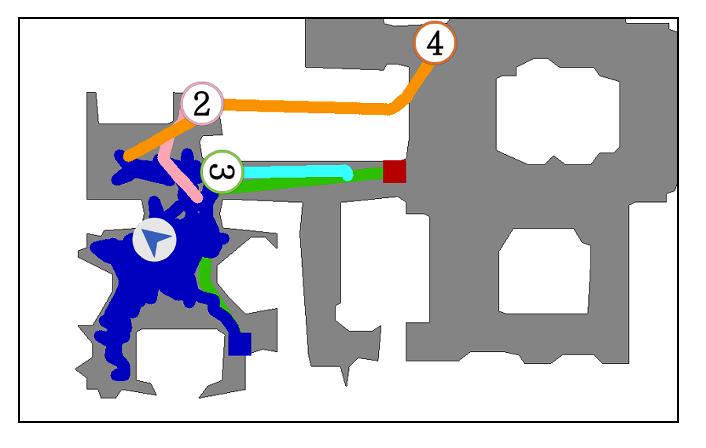}
    \vspace{-2mm}
    \caption{\textbf{Timeout failure example under persistent human blockage.}
The gray region denotes navigable floor and white regions denote non-navigable
space. The green curve is the geodesic path to the red goal, while the dark blue
trajectory shows the robot trajectory. Colored trajectories indicate the
full-episode motion of pedestrians 2--4. A pedestrian repeatedly occupies the
narrow passage along the nominal route, causing Social-WM to remain in the
initial room. The model avoids unsafe motion near the pedestrian, but lacks
long-horizon reasoning to determine whether to
search for an alternative route, eventually resulting in a timeout.}
\label{fig:failure_case}
    \vspace{-4mm}
\end{figure}

\subsection{Limitations and Failure Cases}
\label{sec:limitations}

Although Social-WM substantially reduces human collisions, the overall success
rate improves only modestly because many remaining failures are \emph{timeouts}.
The current world model mainly reasons about short-horizon safety and action
realizability, while long-horizon route selection remains unchanged.

Figure~\ref{fig:failure_case} shows a representative case. A pedestrian
persistently occupies the narrow passage along the nominal route. Social-WM
avoids the unsafe region, but cannot determine whether the blockage will
persist or search for an alternative route, so the episode eventually times
out. Similar failures occur when local decisions require a longer-term
detour.

Image-goal navigation is also challenging over long horizons. The final goal
image provides little guidance when its content is not yet visible, while
route-sampled image subgoals become unreliable after large deviations from the
nominal path. These limitations suggest that future work should combine the
current local safety reasoning with route-level replanning, spatial memory, and
adaptive subgoal selection.

\section{Conclusion}
\label{sec:conclusion}

We presented \textbf{Social-WM}, an efficient latent world-model framework for social navigation that explicitly distinguishes commanded actions from the motion that is actually realizable under physical and social constraints. By combining action-conditioned latent prediction with realizable-action inverse dynamics, Social-WM evaluates candidate action chunks through imagined future transitions before execution. Experiments on Social-HM3D and zero-shot Social-MP3D show that this design substantially reduces human collisions and improves personal-space compliance while maintaining competitive navigation success using only offline RGB trajectories. Our results also reveal that the remaining failures are dominated by long-horizon timeouts, highlighting route-level replanning and adaptive subgoal reasoning as important directions for future work.

\bibliographystyle{IEEEtran}
\bibliography{main}

\end{document}